\documentclass[runningheads]{llncs}

\usepackage{eccv}

\usepackage{eccvabbrv}
\usepackage{graphicx}
\usepackage{booktabs}
\usepackage{soul}
\usepackage{url}
\usepackage[utf8]{inputenc}
\usepackage{graphicx}
\usepackage{amsmath}
\usepackage{xcolor}

\usepackage{adjustbox}
\usepackage{booktabs}
\usepackage{algorithm}
\usepackage{algorithmic}
\usepackage{multirow}
\usepackage{multicol}
\usepackage{dsfont}
\usepackage{amssymb}

\usepackage{pifont}
\usepackage{bbding}
\usepackage{fontawesome}

\newcommand{\cmark}{\ding{51}}
\newcommand{\xmark}{\ding{55}}
\newcommand{\rotbox}[1]{\rotatebox{55}{#1}}

\usepackage[accsupp]{axessibility}  

\usepackage{hyperref}

\usepackage{orcidlink}

\begin{document}
\title{Conceptual Codebook Learning for Vision-Language Models}

%

\author{Yi Zhang\inst{1,2}\orcidlink{0000-0002-5831-0170} \and
Ke Yu\inst{3}\orcidlink{0009-0006-9687-6171} \and
Siqi Wu\inst{4} \and
Zhihai He\inst{2}\orcidlink{0000-0002-2647-8286}\thanks{Corresponding author} }

\authorrunning{Y.~Zhang et al.}

\institute{
Harbin Institute of Technology \and
Southern University of Science and Technology
    \email{zhangyi2021@mail.sustech.edu.cn}
    \email{hezh@sustech.edu.cn}\\ \and
University of California San Diego \\
    \email{key022@ucsd.edu} \and
University of Missouri \\
    \email{siqiwu@missouri.edu}
}

\maketitle

\begin{abstract}
In this paper, we propose Conceptual Codebook Learning (CoCoLe), a novel fine-tuning method for vision-language models (VLMs). CoCoLe aims to address the challenge of enhancing the generalization capability of VLMs while adapting them to downstream tasks in a few-shot setting.
We recognize that visual concepts like shapes, colors, and textures are inherently transferable across different domains and are essential for generalization tasks.
Motivated by this critical finding, we learn a conceptual codebook consisting of visual concepts as keys and conceptual prompts as values, which serves as a link between the image encoder's outputs and the text encoder's inputs. Specifically, for a given image, we leverage the codebook to identify the most relevant conceptual prompts associated with the class embeddings to perform the classification. Additionally, we incorporate a handcrafted concept cache as a regularization to alleviate the overfitting issues in low-shot scenarios. 
This conceptual codebook learning method has been shown to improve the alignment between visual and linguistic modalities.
Extensive experimental results demonstrate that our CoCoLe method remarkably outperforms the existing state-of-the-art methods across various evaluation settings, including base-to-new generalization, cross-dataset evaluation, and domain generalization tasks. Detailed ablation studies further confirm the efficacy of each component in CoCoLe.
\end{abstract}

\section{Introduction}

\begin{figure}[t]
    \centering
    \includegraphics[width=\linewidth]{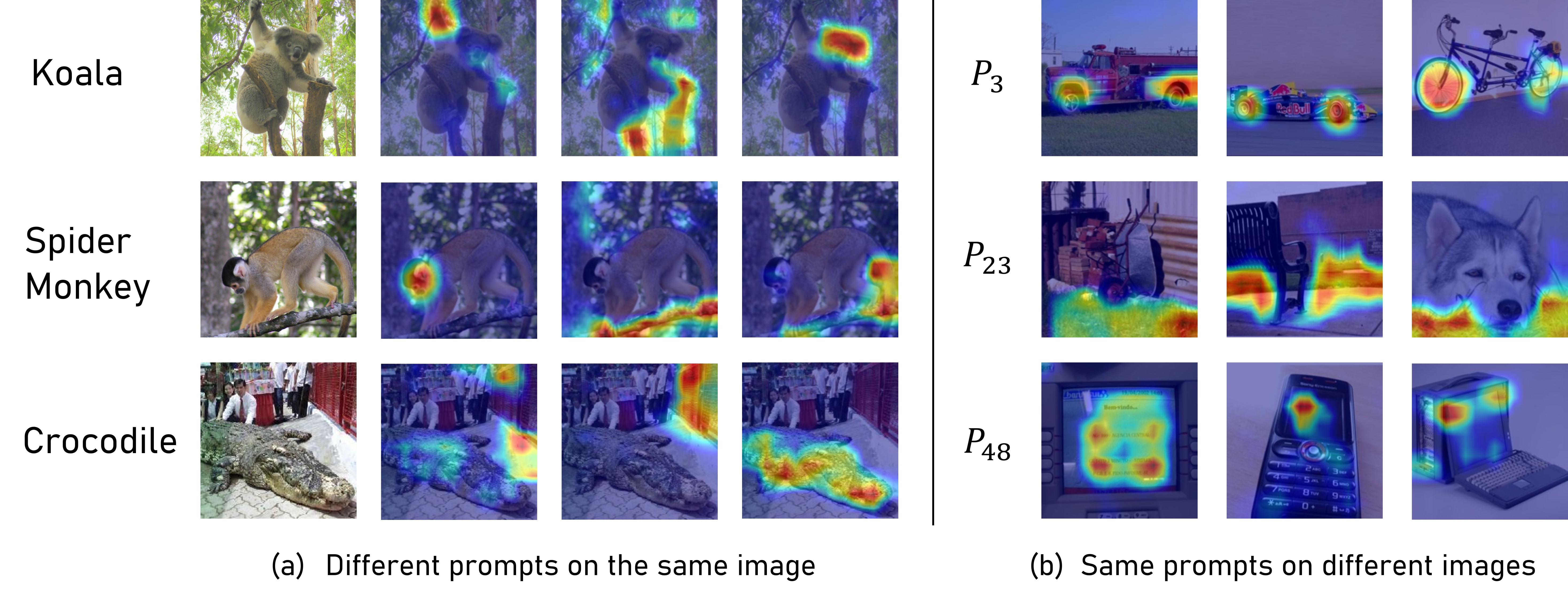}
    \caption{(a) Visualization of the chosen prompts of the same image. (b) Visualization of the same prompts on different images. Grad-CAM~\cite{selvaraju2017grad} is used for the visualization. }
    \label{fig:visual}
\end{figure}

Large-scale pre-trained Vision-Language Models (VLMs), \eg, ALIGN~\cite{jia2021scaling} and CLIP~\cite{radford2021learning}, have achieved exceptional zero-shot performance in various downstream tasks.
These VLMs, trained on massive datasets of image-text pairs with contrastive optimization objectives, effectively align and embed different modalities into a shared vector space. 
Despite their impressive performance, adapting these models to diverse downstream tasks remains challenging due to their substantial size. As a result, recent research has concentrated on improving the ability of pre-trained VLMs to adapt to downstream tasks by fine-tuning supplementary parameters, while keeping VLMs frozen. Prompt-tuning methods, \eg CoOp~\cite{zhou2022learning} and ProGrad~\cite{zhu2022prompt}, replace manual prompts with learnable ones to obtain task-specific knowledge, while adapter-based methods utilize extra modules directly on the top of VLMs, such as Clip-adapter \cite{gao2021clip} and Tip-adapter~\cite{zhang2022tip}. These methods have made significant advancements with limited labeled data.

However, we find that existing fine-tuning methods for CLIP, including CoOp~\cite{zhou2022learning} and CPL~\cite{zhang2024concept}, exhibit relatively low performance on fine-grained datasets like FGVCAircraft \cite{maji2013fine} (aircraft classification), and UCF101 \cite{soomro2012ucf101} (action classification). To address the challenge of enhancing the generalization capability of VLMs in a few-shot settings, in this paper, we propose a novel fine-tuning method called Conceptual Codebook Learning (CoCoLe). Our idea stems from the observation that visual concepts are naturally transferable across domains. As illustrated in \cref{fig:visual}, within a single image, there exist multiple distinct visual concepts focusing on different regions. For example, the selected prompts highlight the claws, ears of the koala, and the branches where the koala stands. Moreover, there are similar concepts in images from different classes; for example, the "firetruck", "racer", and "bicycle" classes possess the compound concept of "wheel" in common.

\begin{figure}[t]
    \centering
    \includegraphics[width=\linewidth]{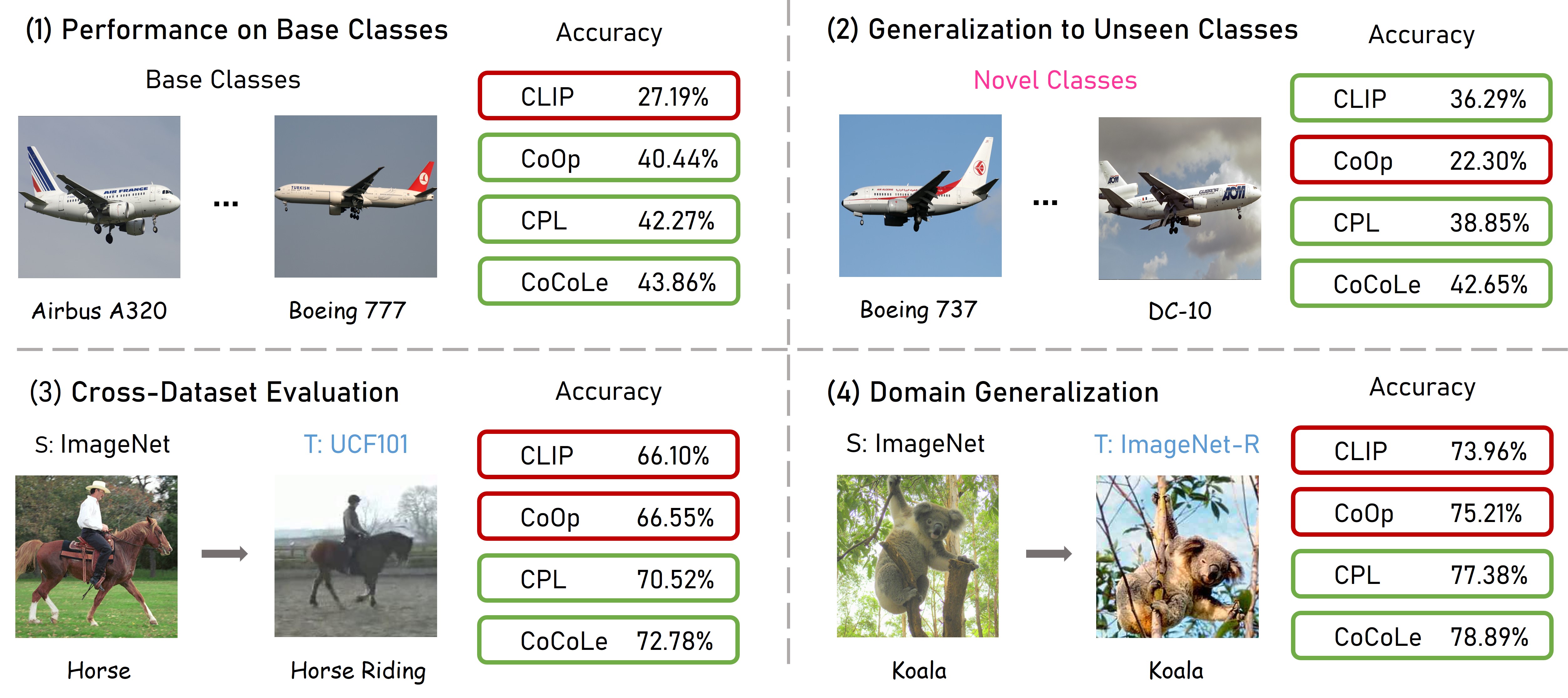}
    \caption{Illustrations and accuracy comparisons on base-to-novel generalization, cross-dataset transfer and domain generalization tasks. S and T represent source and target datasets respectively. }
    \label{fig:intro}
    
\end{figure}

Motivated by this interesting finding, we propose to learn a conceptual codebook consisting of visual concepts as keys and conceptual prompts as values, which serve as a link between the image encoder's outputs and the text encoder's inputs. Specifically, for a given image, we leverage the codebook to identify the most relevant conceptual prompts associated with the class embeddings to perform the classification. Additionally, we incorporate a handcrafted concept cache as a regularization to alleviate the overfitting issues in low-shot scenarios. As shown in \cref{fig:intro}, we observe that this conceptual codebook learning method can achieve enhanced alignment between visual and linguistic modalities. Our contributions could be summarized as: 
\begin{itemize}
    \item We proposed a novel fine-tuning method named CoCoLe for VLMs to solve the problem of performance degradation on generalization tasks.
    \item CoCoLe introduces a conceptual codebook to adaptively learn visual concepts and their corresponding conceptual prompts with regularization to further guarantee the generalization capability.
    \item Extensive experimental results demonstrate the outstanding performance of CoCoLe compared to existing state-of-the-art methods in base-to-novel generalization, domain generalization, and cross-dataset transfer tasks.
\end{itemize}

\section{Related Work}

\subsection{Vision-Language Models}
Pre-trained vision-language models (VLMs) have recently emerged as a notable trend~\cite{lei2015predicting,sariyildiz2020learning,desai2021virtex,radford2021learning}. Capitalizing on tremendous image-text data, these large-scale models can effectively acquire visual representations using contrastive loss, enabling them to grasp both visual and textual semantics and achieve successful modality alignment. Current studies~\cite{zhou2022learning,zhang2021vt} have showcased that by harnessing extensive sets of image-text pairs, VLMs exhibit outstanding performance across a range of downstream visual tasks~\cite{hu2023learning,duan2022multi,kan2023knowledge}. For example, Derived through contrastive learning on 400 million online image-text pairs, CLIP \cite{radford2021learning} demonstrated remarkable zero-shot accuracy on classification tasks. Our method aims to utilize the comprehensive capability of CLIP to perform knowledge-guide fine-tuning for better adaptation to downstream tasks.

\subsection{Prompt Tuning for VLMs}
As text input for pre-trained vision-language models, prompts function as the guidance for the downstream tasks, extracting task-specific information from the existing knowledge within VLMs~\cite{zhou2022conditional,zhou2022learning}. Setting a precedent in this field, CoOp~\cite{zhou2022learning} exploits a set of learnable vectors to perform end-to-end optimization on the prompt context but fails to generalize to unseen classes. To address this issue, CoCoOp~\cite{zhou2022conditional} improved CoOp's generalization by generating conditional prompts.
Further, KgCoOp~\cite{yao2023visual} enhances the generalization by minimizing the discrepancy between learned and handcrafted prompts, and CoPrompt~\cite{roy2024consistency} constrains the trainable models by pre-trained ones to avoid the overfitting problem on the downstream task. Meanwhile, there are methods exploring diverse forms of prompts. MaPLe~\cite{khattak2023maple} enhances both vision and language components by employing a coupling function to promote cross-modal synergy, whereas CPL~\cite{zhang2024concept} leverages the powerful generalization of CLIP to build a visual concept cache with a projector to capture multi-level visual features.

In our work, we mainly focus on prompt-tuning ways and meticulously manipulate learnable vectors by a learnable codebook with the regularization of a handcrafted concept cache. Among existing methods, the most related to ours are CoOp and CPL. 
Compared with CoOp, the proposed CoCoLe introduces an adaptive codebook rather than fixed to specific classes or tasks. On the other hand, CoCoLe leverages the transferability of concepts across domains, with optimal handcrafted concept-based prompts as a regularization to prevent the codebook from overfitting.

\subsection{Visual Concept Learning}
Earlier studies have identified two primary methods for visual concept learning. The first approach generally involves using manual annotations of concepts (e.g., textures, fabrics, and colors) for the training images~\cite{patterson2012sun,patterson2016coco}, while the other method utilizes unsupervised learning to design data-driven concepts~\cite{fei2005bayesian,liu2011recognizing,huang2016unsupervised}.  However, these approaches may introduce biases into the learned concepts, limiting their overall performance. 
Recent studies have sought complementary prompting methods to capture unbiased visual concepts \cite{wang2022dualprompt,wang2023attriclip}. Notably, CPL~\cite{zhang2024concept} first utilizes the capabilities of CLIP~\cite{radford2021learning} to design an unsupervised concept cache. Furthermore, in this work, we introduce a learnable codebook supervised by a handcrafted concept cache, which automatically selects conceptual prompts that are aligned with visual concepts.

\section{Method}

\begin{figure}[t]
    \centering
    \includegraphics[width=\linewidth]{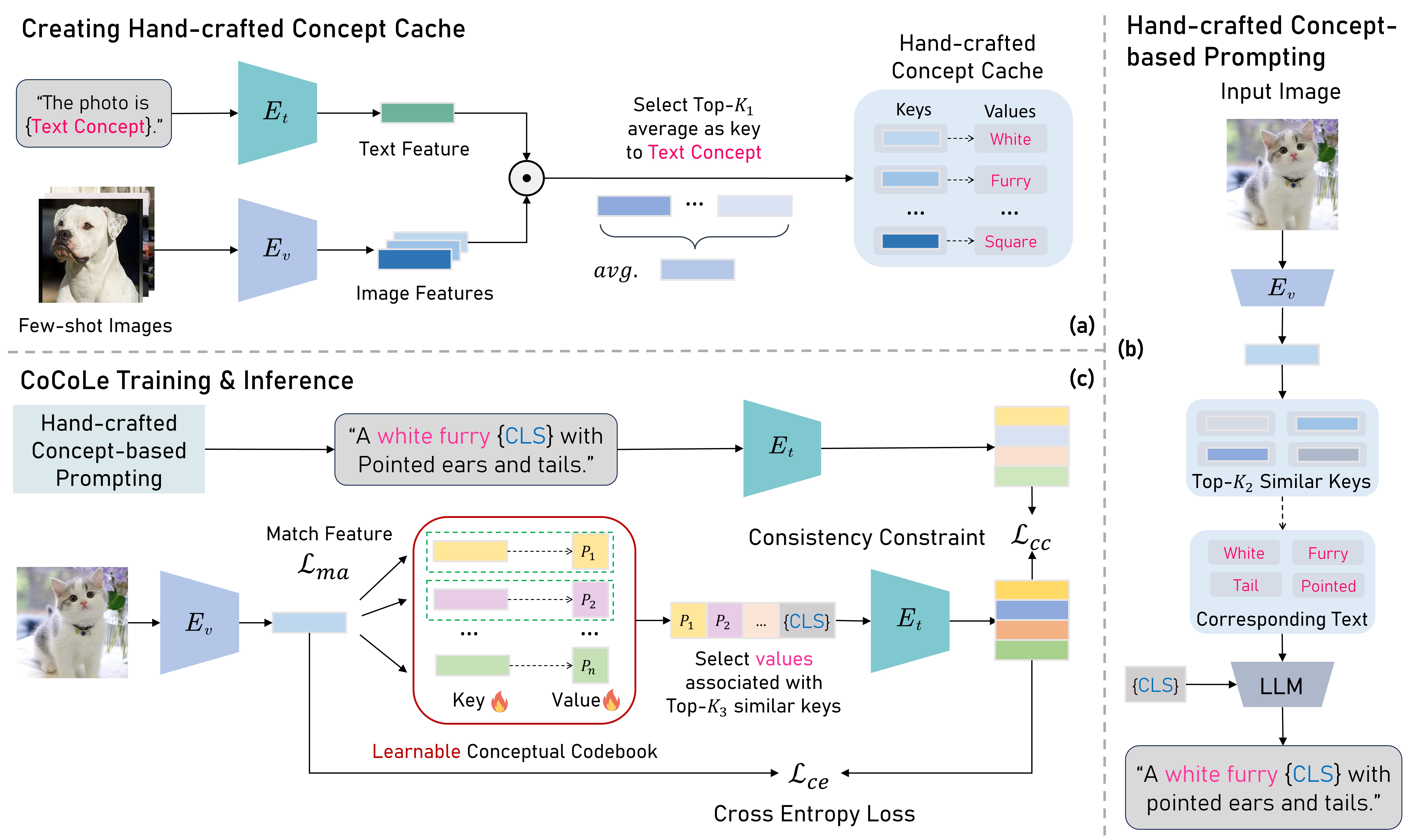}
    \caption{An overview of the proposed CoCoLe. (a) shows the establishing process of handcrafted concept cache. (b) displays the handcrafted concept-based prompting process. (c) presents the training pipeline for CoCoLe. Within CoCoLe, only the keys and values in the Conceptual Codebook are learnable.}
    \label{fig:concept_ov}
\end{figure}

\subsection{Background and Overview}
\label{sec:background}
\paragraph{\textbf{CLIP and CoOp.}} CLIP~\cite{radford2021learning} consists of two main encoders: the visual encoder $E_v$ for processing image inputs $X_v$, and the text encoder $E_t$ for handling textual prompts $T_c$, which are formatted as ``\texttt{a photo of} $[\text{CLS}]_c$", where $[\text{CLS}]_c$ is the word embedding for class $c$. CLIP optimizes the similarity between the image features and the prompt embeddings that correspond to the true labels during training.
CoOp~\cite{zhou2022learning} takes this a step further by replacing manually constructed prompts with learnable ones. It employs a set of $n$ adaptable context vectors ${[V_{1}], [V_{2}],\cdots , [V_{n}]}$, each with the same dimension as word embeddings. Gradient descent is utilized to optimize the learnable vectors. For a given class $c$, the corresponding prompt is formulated as $T_{c} = {[V_{1}], [V_{2}],\cdots , [V_{n}], [\text{CLS}]_c}$.

\paragraph{\textbf{Overview of CoCoLe. }}In \cref{fig:concept_ov}, we illustrate an overview of our proposed CoCoLe approach. \Cref{fig:concept_ov} (a) presents the process of constructing the handcrafted concept cache. We begin by compiling a list of text concepts $\Omega_t$ that encapsulates key visual concepts. Next, we utilize CLIP's powerful image-text association capability to identify the image feature $v_j$ that has the Top-$K_1$ highest similarity scores for each text concept feature $c_t^i \in C_t$. The highest scoring Top-$K_1$ features are averaged and saved as keys in the visual concepts cache, paired with their respective text concepts $\omega_i \in \Omega_t$ as values.
In \Cref{fig:concept_ov} (b), we illustrate the handcrafted concept-based prompting process: initially, we use $E_v$ to obtain the image feature $v$. This image feature is then used as a query to identify the Top-$K_2$ similar keys based on cosine distance. Subsequently, the associated values, combined with the class embeddings, serve as input for LLM (\eg, GPT~\cite{brown2020language}) to generate the optimal handcrafted concept-based prompts represented as $\mathcal{P}^h \triangleq \{ P_{h_i}\}_{i=1}^{N^C}$, where $N^C$ denotes the number of class.

\Cref{fig:concept_ov} (c) shows the CoCoLe training pipeline. We begin by using the visual encoder to obtain the visual features $f_v$ from a given image $x$.
We then follow (b) to generate the handcrafted concept-based prompts $\mathcal{P}^h$
and obtain text features by $E_t$, denoted as $\mathcal{F}_h = E_t(\mathcal{P}^h)$. 
The visual concepts (keys) and the conceptual prompts (values) in the conceptual codebook are trainable parameters optimized by four loss functions. The classification loss $\mathcal{L}_{ce}$ is employed to maximize the alignment between the image feature $f_v$ and the related text features $f_t$. We utilize the loss function $\mathcal{L}_{ma}$ to minimize the distance between the chosen keys (Top-$K_3$ similar visual concepts) and the image feature $f_v$, facilitating the learning of generalizable concepts by the keys.
$\mathcal{L}_{cc}$ works as a regularization for diminishing the overfitting problem, ensuring the text features produced by the selected learned prompts do not deviate significantly from those generated by the handcrafted concept-based prompts.
Finally, $\mathcal{L}_{or}$ ensures that the text features of the prompts are orthogonal to enhance the prompts' diversity.

\subsection{Conceptual Codebook Learning (CoCoLe)}
\label{sec:ccl}
\paragraph{\textbf{Learnable Conceptual Codebook. }}In the CoOp framework, each class embedding is associated with a single set of prompt vectors. Nevertheless, images belonging to the same class often encompass a variety of concepts. Conflating these varied concepts into a single set of prompts can result in significant knowledge loss. Moreover, the encoded information within CoOp's prompts lacks inter-class interaction, since concepts from one class may assist in identifying another class with similar concepts. For instance, when presented with an image of a cat in the tree, the concept of "in the tree" might also apply to images of other animals (e.g., a koala in the tree). We hypothesize that fine-tuning prompts based on image concepts can facilitate the learning of textual descriptions associated with these concepts, thereby improving generalization across datasets.

As such, we propose CoCoLe, as depicted in \cref{fig:concept_ov}. The key insight of CoCoLe is a trainable concept codebook, empowering the image to autonomously determine the prompts it should learn based on its inherent concepts. For each training input, only a subset of prompts that align with the current image concepts are chosen and trained individually.
The learnable concept codebook stores visual concepts as keys and conceptual prompts as values, comprising $N$ (key, value) pairs, denoted as $\Psi_{cc} \triangleq \{(\mathbf{V}_i, \mathbf{P}_i)\}_{i=1}^N$,
where $\Psi_{cc}$ denotes the {learnable concept codebook}, each $\mathbf{V}_i\!\in\!\mathbb{R}^D$ shares the same dimensionality as the image feature $f_v$. Additionally, each $\mathbf{P}_i\!=\![\mathbf{p}_i]_1 \ldots [\mathbf{p}_i]_M\!\in\!\mathbb{R}^{D\times M}$ consists of $M$ learnable vectors.
We represent the set of learnable visual concepts as $\mathcal{V}\!=\!\{\mathbf{V}_i\}_{i=1}^N$ and the entire set of  learnable conceptual prompts as $\mathcal{P}\!=\!\{\mathbf{P}_i\}_{i=1}^N$. 
In an optimal scenario, we anticipate that the image itself should determine the prompts to be selected, guided by the concepts it encompasses, in order to steer the prediction process. 
To achieve this, for an input image $\mathbf{x}_j$, we first extract its image feature $f_{v_j}\!=\!E_v(\mathbf{x}_j)$, where $j$ represents image index.
Then we calculate the cosine similarity score between $f_{v_j}$ and $\mathbf{V}_i \in \mathcal{V}$, denoted as, $S_c=\frac{f_{v_j} \cdot \mathbf{V}_i}{||f_{v_j} || ~ ||\mathbf{V}_i||}$.

Next, we choose the keys with Top-$K_3$ cosine similarity score to form set $\mathcal{V}_j$, representing the subset of Top-$K_3$ visual concepts chosen from $\mathcal{V}$ uniquely for the $j$-th image.
Then, we select the conceptual prompts that match these visual concepts, represented as $\mathcal{P}_j\!=\!\{\mathbf{P}_{j_i}\}_{i=1}^{K_3}$, {where $\mathbf{P}_{j_i}$ denotes the $i$-th prompt chosen uniquely for $\mathbf{x}_j$.} We use 
these prompts to link to the class name embedding of $\mathbf{x}_j$ as shown in Fig.~\ref{fig:concept_ov}, and the input for text encoder can be represented as, $\mathbf{T}(\mathcal{P}_j)=\operatorname{concat}(\mathbf{P}_{j_1};\dots;\mathbf{P}_{j_{K_3}};[\text{CLS}]_d), $
where $\operatorname{concat}(\cdot)$ signifies concatenation. 
Thus, for a test image $\mathbf{x}_j$ and prompts $\mathcal{P}_j$ based on the concepts of $\mathbf{x}_j$, the text feature $f_{t_j}$ can be obtained by $f_{t_j} \triangleq E_t(\mathbf{T}(\mathcal{P}_j))$.
The likelihood of predicting the image as class $y_i$ is ultimately determined by:
\begin{equation}
\textbf{}
    p(y_i|\mathbf{x}_j)=\frac{e^{\langle f_{v_j},f_{t_j}\rangle/\tau}}{\sum_{d=1}^D e^{\langle f_{v_j}, f_{t_d}\rangle/\tau}}.
    \textbf{}
    \label{eq:our method}
\end{equation}

From a broader viewpoint, the suggested {adaptable concept codebook} serves as a link connecting the outcomes of the image encoder and the inputs of the text encoder. The keys are fine-tuned to closely align with the identified image features, which hold abundant high-level information, such as image concepts. Meanwhile, the prompts are refined to encompass textual details associated with the respective image concepts, facilitating improved guidance for the model predictions alongside the class name embeddings.

\begin{figure}[t]
    \centering
    \includegraphics[width=\linewidth]{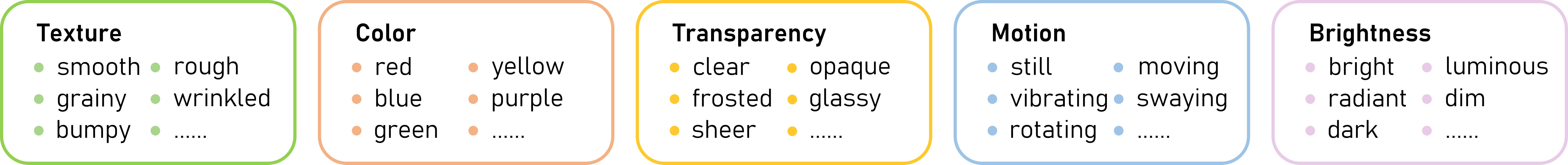}
    \caption{Examples of text concepts from established visual concept datasets, including descriptive terms of texture, color, transparency, motion and brightness.}
    \label{fig:concept_cache}
\end{figure}

\paragraph{\textbf{Handcrafted Concept Cache.}} 
In Figure \ref{fig:concept_ov} (a), inspired by~\cite{zhang2023cross}, we construct a comprehensive list $\Omega_t$ containing $I=2000$ descriptive text concepts sourced from established visual concept datasets~\cite{zhang2024concept,zhang2023cross}. These descriptions encompass terms related to texture, colors, brightness, density, etc., categorized into 50 classes. Examples of these terms are depicted in \cref{fig:concept_cache}. The dictionary is defined as $\Omega_t \triangleq \{\omega_i\}_{i=1}^I$. Following CLIP's zero-shot setup, we start by appending each $\omega_i$ to a manually crafted prompt $\phi =$ ``\texttt{The photo is} ..." to create a concept-specific textual input $\{\pi; \omega_i\}$. Subsequently, using the text encoder $E_t$, we derive text concept features $C_t \triangleq \{ c_t^i \}_{i=1}^I$, where each $c_t^i = E_t({\pi; \omega_i})$.

We denote the handcrafted concept cache as $\Phi_{mc} \triangleq \{(key, value)_i\}_{i=1}^I$. The key and value are the visual concepts and textual concept words respectively. 
The visual concepts are identified by exploiting the text concept features $C_t$ extracted from the CLIP model, using training images as a basis. 
In the context of $H$-shot $D$-class few-shot learning, there are $H$ labeled images available for each of the $D$ classes. Using the CLIP visual encoder $E_v$, we obtain their corresponding image features $V \triangleq \{ v_j \}_{j=1}^{HD}$, where each $v_j = E_v(x_j)$. Each text concept feature $c_t \in C_t$ is matched against all visual features in $V$  using similarity score formula $S_t = \mathrm{sim}\left(c_t,v_j \right)= c_tv_j$, where both $c_t$ and $v_j$ are normalized. Afterwards, we select the Top-$K_1$ image features with the highest similarity scores, computing their average as the key, and associating it with the corresponding text concept word $\omega_t$. These pairs are stored within the handcrafted concept cache.

\paragraph{\textbf{Conceptual Codebook Learning with Regularization.}} 
\Cref{fig:concept_ov} (b) presents the handcrafted concept-based prompting process. Initially, we extract the image feature $f_v$ using $E_v$. Subsequently, we employ this image feature as the query to retrieve the Top-$K_2$ most similar keys based on cosine similarity. Ultimately, we obtain the corresponding values (conceptual words).
Together with the class name, these concept words are input to an LLM (eg. GPT~\cite{brown2020language}) to generate optimal handcrafted concept-based prompts. 
Therefore, our approach addresses the challenge of diminished generalization on downstream tasks by introducing a regularization, ensuring the text features produced by the selected learned prompts do not differ significantly from their counterpart generated by the handcrafted concept-based prompts.
We enforce this consistency by utilizing the Euclidean distance as a constraint between the text features generated from the hand-crafted concept-based prompts ($f_{t_d}^h$) and those obtained from selected learned prompts ($f_{t_d}^l$). While alternative measures such as cosine distance could also serve as constraints, our empirical findings suggest that Euclidean distance yields superior performance. We can represent the consistency constraint as:
\begin{equation}
\mathcal{L}_{cc}=\frac{1}{D}\sum_{d=1}^{D}||f_{t_d}^l-f_{t_d}^h||^{2}_{2},
\end{equation}
where $||\cdot||$ is the euclidean distance, $D$ is the number of seen classes.

\subsection{{Training and Inference }}
\label{sec3.3}
\paragraph{\textbf{Training Objective.}}
According to Equation~\ref{eq:our method}, the loss for image classification is expressed as:
\begin{equation}
    \mathcal{L}_{ce}=\mathbb{E}[-\text{log}\frac{e^{\langle f_{v_j},f_{t_j}\rangle/\tau}}{\sum_{d=1}^D e^{\langle f_{v_j},f_{t_d}\rangle/\tau}}].
    \label{eq:Lm}
\end{equation}

Besides $\mathcal{L}_{ce}$, we require a matching loss that brings the matched top-$K_3$ keys $\mathcal{K}_j$ closer to the image embedding $\mathbf{z}_j$, facilitating the keys to learn diverse concepts from the samples. We employ cosine distance as our matching loss. Nevertheless, alternative metrics such as Euclidean distance can also serve as a viable matching loss. Through empirical observation, we find that cosine distance typically delivers optimal performance as a matching loss. Therefore,
The matching loss adopted to optimize the keys is defined as:
\begin{equation}
\label{eq:match}
    \mathcal{L}_{ma} = \sum_{i=1}^C (1 - \frac{f_{v_j} \cdot \mathbf{V}_{j_i}}{||f_{v_j} || ~ ||\mathbf{V}_{j_i}||}). 
\end{equation}

Lastly, to enhance the semantic diversity of the learned prompts, we introduce an additional loss that orthogonalizes the embeddings of different prompts, thereby boosting prompt diversity.

\begin{equation}
\mathcal{L}_{or}=\frac{1}{N(N-1)}\sum_{i=1}^{N}\sum_{j=i+1}^N|\langle E_t(\mathbf{P}_i), E_t(\mathbf{P}_j)\rangle|,
\label{eq:Lp}
\end{equation}
where $\langle\cdot,\cdot\rangle$ denotes the cosine similarity. In this way, the overall optimization objective is defined as:
\begin{equation}    \mathcal{L}=\mathcal{L}_{ce}+\mathcal{L}_{ma}+\mathcal{L}_{or}+\mathcal{L}_{cc},
\label{eq:total loss}
\end{equation}
The keys in the conceptual codebook are optimized by $\mathcal{L}_{ma}$, while the prompts are optimized through $\mathcal{L}_{ce}$, $\mathcal{L}_{or}$ and $\mathcal{L}_{cc}$.

 \paragraph{\textbf{Inference.}} Once visual and textual concepts (keys and values) are learned through training, they can be shipped with CLIP for downstream tasks with a standard zero-shot CLIP inference setup. As shown in \Cref{fig:concept_ov}, we first generate the visual features $f_{vt}$ of the test image using visual Encoder $E_v$, then we use $f_{vt}$ to choose top-$K_3$ similar visual concepts(keys) by cosine similarity, next, we concatenate the corresponding prompts(values) of the keys to obtain the conceptual prompt, which is fused with each given name to produce conceptual prompt text features $\{f_{tt}^i\}_{i=1}^C$. Finally, the zero-shot inference is performed with the conceptual prompted text features and the input image feature $f_{vt}$ to produce classification scores on the test images.

\section{Experiments}

\subsection{Experimental Setup}
\paragraph{\textbf{Benchmark Settings.}} We follow previous works to extensively evaluate our proposed method on three challenging tasks:
\begin{itemize}
    \item \textbf{Base to Novel Class generalization.} We evaluate the generalization capability of our method in zero-shot scenarios within a dataset. The dataset is evenly divided into base and novel classes. We train our model with few-shot images on the base classes and then do the evaluation on both base classes and unseen novel classes.
    \item \textbf{Cross-Dataset Evaluation.} The cross-dataset transfer is a much more challenging generalization task compared to base-to-novel generalization, since the latter only transfers within a single dataset while the former transfers across different datasets, \eg, from object recognition to texture classification. In this experiment,
    we follow previous works to train our model in a few-shot setting on 1000 ImageNet classes and subsequently evaluate its performance on ten other unseen datasets.
    \item \textbf{Domain Generalization.} 
    We assess the performance of our model on out-of-distribution generalization. Likewise, we evaluate our model trained on ImageNet directly on four variants of ImageNet, each containing the same classes but from different distributions.
\end{itemize}

\paragraph{\textbf{Datasets.}} 

For conducting experiments on base-to-novel generalization and cross-dataset transfer tasks, we adhere to the setting of prior studies~\cite{radford2021learning,zhou2022learning,zhou2022conditional}. Specifically, we evaluate our approach across 11 diverse image classification datasets. These datasets encompass a wide range of tasks, including generic object classification (e.g., ImageNet~\cite{deng2009imagenet} and Caltech101~\cite{fei2004learning}), fine-grained classification (e.g., OxfordPets~\cite{parkhi2012cats}, StanfordCars~\cite{krause20133d}, Flowers102~\cite{nilsback2008automated}, Food101~\cite{bossard2014food}, and FGVCAircraft~\cite{maji2013fine}), scene recognition (SUN397~\cite{xiao2010sun}), action recognition (UCF101~\cite{soomro2012ucf101}), texture classification (DTD~\cite{cimpoi2014describing}), and satellite image recognition (EuroSAT~\cite{helber2019eurosat}). For the domain generalization task, we employ ImageNet as the source dataset and evaluate our method's performance on four ImageNet variants: ImageNet-A~\cite{hendrycks2021natural}, ImageNet-R~\cite{hendrycks2021many}, ImageNet-V2~\cite{recht2019imagenet}, and ImageNet-Sketch~\cite{wang2019learning}.

\begin{table*}[t] 
    \setlength{\tabcolsep}{1.6pt}
    \caption{\small{Comparison with existing methods on base-to-novel generalization}. The top accuracies are highlighted in bold, with the second-best results underlined. The harmonic mean is denoted by HM. }
    \label{table:b_to_n}
    \begin{subtable}[t]{.243\textwidth}
    \centering
    \caption{\textbf{Average}}
    \resizebox{0.99\linewidth}{!}{
    \label{table:average_acc}
    \begin{tabular}{l cc|c}
    \toprule
    & Base & Novel & HM \\
    \midrule
    CLIP & 69.34 & 74.22 & 71.70 \\
    CoOp &  {82.69} & 63.22 & 71.66 \\
    Co-CoOp & 80.47 & 71.69 & 75.83 \\
    KgCoOp & 80.73 & 73.60 & 77.00 \\ 
    MaPLe & 82.28 &  {75.14} &  {78.55} \\
    CoPrompt & 84.00 & 77.23 & 80.48 \\
    CPL & \underline{84.38} & \underline{78.03} & \underline{81.08} \\
    \midrule
    Ours & \textbf{85.22} & \textbf{80.31} & \textbf{82.70} \\
     & \textcolor{blue}{+0.84} & \textcolor{blue}{+2.28} & \textcolor{blue}{+1.62} \\
     
    \bottomrule
    \end{tabular}
    }
    \end{subtable}
    \begin{subtable}[t]{.243\textwidth}
    \centering
    \caption{ImageNet}
    \resizebox{0.99\linewidth}{!}{
    \begin{tabular}{l cc|c}
    \toprule
    & Base & Novel & HM \\
    \midrule
    CLIP & 72.43 & 68.14 & 70.22 \\
    CoOp & {76.47} & 67.88 & 71.92\\
    Co-CoOp & 75.98 & {70.43} & {73.10} \\
    KgCoOp &75.83 &69.96& 72.78\\
    MaPLe & {76.66} &  {70.54} & {73.47} \\
    CoPrompt & 77.67 &	 71.27 &	 74.33 \\
    CPL & \underline{78.74} & \underline{72.03} & \underline{75.24} \\
    \midrule
    Ours & \textbf{79.25} &	 \textbf{74.58} &	 \textbf{76.84} \\ 
         & \textcolor{blue}{+0.51} & \textcolor{blue}{+2.55} & \textcolor{blue}{+1.60} \\
    \bottomrule
    \end{tabular}
    }
    \end{subtable}
    \begin{subtable}[t]{.243\textwidth}
    \centering
    \caption{Caltech101}
    \resizebox{0.99\linewidth}{!}{
    \begin{tabular}{l cc|c}
    \toprule
    & Base & Novel & HM \\
    \midrule
    CLIP & 96.84 & {94.00} & 95.40 \\
    CoOp & {98.00} & 89.81 & 93.73 \\
    Co-CoOp & 97.96 & 93.81 & {95.84} \\
    KgCoOp  & 97.72  & 94.39  & 96.03 \\ 
    MaPLe & 97.74 & {94.36} &  {96.02} \\
    CoPrompt & \underline{98.27} & 94.90 & 96.55 \\
    CPL & \textbf{98.35} & \underline{95.13} & \underline{96.71} \\ 
    \midrule
    Ours & 98.17 & \textbf{95.67} & \textbf{96.90} \\
         & \textcolor{red}{-0.18} & \textcolor{blue}{+0.54} & \textcolor{blue}{+0.19} \\
    \bottomrule
    \end{tabular}
    }
    \end{subtable}
    \begin{subtable}[t]{.243\textwidth}
    \centering
    \caption{OxfordPets}
        \resizebox{0.99\linewidth}{!}{     \begin{tabular}{l cc|c}
    \toprule
    & Base & Novel & HM \\
    \midrule
    CLIP & 91.17 & 97.26 & 94.12 \\
    CoOp & 93.67 & 95.29 & 94.47 \\
    Co-CoOp & {95.20} & {97.69} & {96.43} \\
    KgCoOp & 94.65 & 97.76 & 96.18 \\ 
    MaPLe &  {95.43} & {97.76} &  {96.58} \\
    CoPrompt & 95.67 & 98.10 & 96.87 \\
    CPL & \underline{95.86} & \underline{98.21} & \underline{97.02} \\ 
    \midrule
    Ours & \textbf{96.21} & \textbf{98.55} & \textbf{97.37} \\ 
         & \textcolor{blue}{+0.35} & \textcolor{blue}{+0.34} & \textcolor{blue}{+0.35} \\
    \bottomrule
    \end{tabular}
    }
    \end{subtable}
    \\
    \begin{subtable}[t]{.243\textwidth}
    \centering
    \caption{StanfordCars}
        \resizebox{0.99\linewidth}{!}{     \begin{tabular}{l cc|c}
    \toprule
    & Base & Novel & HM \\
    \midrule
    CLIP & 63.37 &  {74.89} & 68.65 \\
    CoOp & {78.12} & 60.40 & 68.13 \\
    Co-CoOp & 70.49 & 73.59 & {72.01} \\  
    KgCoOp  & 71.76  & 75.04  & 73.36 \\ 
    MaPLe & 72.94 & 74.00 &  {73.47} \\
    CoPrompt & 76.97 & 74.40 & 75.66 \\ 
    CPL & \underline{79.31} & \underline{76.65} & \underline{77.96} \\
    \midrule
    Ours & \textbf{80.32} & \textbf{78.84} & \textbf{79.57} \\ 
         & \textcolor{blue}{+1.01} & \textcolor{blue}{+2.19} & \textcolor{blue}{+1.61} \\
    \bottomrule
    \end{tabular}
    }
    \end{subtable}
    \begin{subtable}[t]{.243\textwidth}
    \centering
    \caption{Flowers102}
        \resizebox{0.99\linewidth}{!}{     \begin{tabular}{l cc|c}
    \toprule
    & Base & Novel & HM \\
    \midrule
    CLIP & 72.08 & 77.80 & 74.83 \\
    CoOp & {97.60} & 59.67 & 74.06 \\
    Co-CoOp & 94.87 & 71.75 & {81.71} \\  
    KgCoOp  & 95.00  & 74.73  & 83.65 \\ 
    MaPLe & 95.92 & 72.46 &  {82.56} \\
    CoPrompt & {97.27} & {76.60} & {85.71} \\ 
    CPL & \textbf{98.07} & \underline{80.43} & \underline{88.38} \\
    \midrule
    Ours & \underline{97.72} & \textbf{81.04} & \textbf{88.60} \\ 
         & \textcolor{red}{-0.35} & \textcolor{blue}{+0.61} & \textcolor{blue}{+0.22} \\
    \bottomrule
    \end{tabular}
    }
    \end{subtable}
    \begin{subtable}[t]{.243\textwidth}
    \centering
    \caption{Food101}
        \resizebox{0.99\linewidth}{!}{      \begin{tabular}{l cc|c}
    \toprule
    & Base & Novel & HM \\
    \midrule
    CLIP & 90.10 & 91.22 & 90.66 \\
    CoOp & 88.33 & 82.26 & 85.19 \\
    Co-CoOp & 90.70 & 91.29 & 90.99 \\ 
    KgCoOp & 90.50 &  91.70 & 91.09 \\
    MaPLe &  90.71 &  92.05 &  91.38 \\
    CoPrompt & 90.73 & 92.07 & 91.40\\
    CPL & \underline{91.92} & \underline{93.87} & \underline{92.88} \\ 
    \midrule
    Ours & \textbf{92.23} & \textbf{94.28} & \textbf{93.24} \\
        & \textcolor{blue}{+0.31} & \textcolor{blue}{+0.41} & \textcolor{blue}{+0.36} \\
    \bottomrule
    \end{tabular}
    }
    \end{subtable}
    \begin{subtable}[t]{.243\textwidth}
    \centering
    \caption{FGVCAircraft}
        \resizebox{0.99\linewidth}{!}{     \begin{tabular}{l cc|c}
    \toprule
    & Base & Novel & HM \\
    \midrule
    CLIP & 27.19 &  {36.29} & {31.09} \\
    CoOp & {40.44} & 22.30 & 28.75 \\
    Co-CoOp & 33.41 & 23.71 & 27.74 \\ 
    KgCoOp & 36.21 & 33.55 & 34.83 \\
    MaPLe & 37.44 & 35.61 & {36.50} \\
    CoPrompt & {40.20} & \underline{39.33} & {39.76} \\
    CPL & \underline{42.27} & 38.85 & \underline{40.49} \\
    \midrule
    Ours & \textbf{43.86} & \textbf{42.65} & \textbf{43.25} \\ 
    & \textcolor{blue}{+1.59} & \textcolor{blue}{+3.32} & \textcolor{blue}{+2.76} \\
    \bottomrule
    \end{tabular}
    }
    \end{subtable}
    \\
    \begin{subtable}[t]{.243\textwidth}
    \centering
    \caption{SUN397}
        \resizebox{0.99\linewidth}{!}{     \begin{tabular}{l cc|c}
    \toprule
    & Base & Novel & HM \\
    \midrule
    CLIP & 69.36 & 75.35 & 72.23 \\
    CoOp & {80.60} & 65.89 & 72.51 \\
    Co-CoOp & 79.74 & {76.86} & {78.27} \\ 
    KgCoOp & 80.29 & 76.53 & 78.36 \\
    MaPLe & {80.82} &  {78.70} &  {79.75} \\
    CoPrompt & \underline{82.63} & \underline{80.03} & \underline{81.31} \\
    CPL & 81.88 & 79.65 & 80.75 \\ 
    \midrule
    Ours & \textbf{83.97} & \textbf{82.24} & \textbf{83.10} \\ 
    & \textcolor{blue}{+1.34} & \textcolor{blue}{+2.21} & \textcolor{blue}{+1.79} \\
    \bottomrule
    \end{tabular}
    }
    \end{subtable}
    \begin{subtable}[t]{.243\textwidth}
    \centering
    \caption{DTD}
    \resizebox{0.99\linewidth}{!}{     \begin{tabular}{l cc|c}
    \toprule
    & Base & Novel & HM \\
    \midrule
    CLIP & 53.24 & {59.90} & 56.37 \\
    CoOp & {79.44} & 41.18 & 54.24 \\
    Co-CoOp & 77.01 & 56.00 & {64.85} \\ 
    KgCoOp & 77.55 & 54.99 & 64.35 \\ 
    MaPLe & {80.36} & 59.18 &  {68.16} \\
    CoPrompt & \textbf{83.13} & \underline{64.73} & \underline{72.79} \\
    CPL & 80.92 & 62.27 & 70.38 \\ 
    \midrule
    Ours & \underline{82.46} & \textbf{68.38} & \textbf{74.76} \\ 
    & \textcolor{red}{-0.67} & \textcolor{blue}{+3.65} & \textcolor{blue}{+1.97} \\
    \bottomrule
    \end{tabular}
    }
    \end{subtable}
    \begin{subtable}[t]{.243\textwidth}
    \centering
    \caption{EuroSAT}
    \resizebox{0.99\linewidth}{!}{     \begin{tabular}{l cc|c}
    \toprule
    & Base & Novel & HM \\
    \midrule
    CLIP & 56.48 & {64.05} & 60.03 \\
    CoOp & {92.19} & 54.74 & 68.69 \\
    Co-CoOp & 87.49 & 60.04 & {71.21} \\  
    KgCoOp & 85.64 & 64.34 & 73.48 \\ 
    MaPLe & {94.07} &  {73.23} & {82.35} \\
    CoPrompt & \underline{94.60} & 78.57 & 85.84 \\
    CPL & 94.18 & \underline{81.05} & \underline{87.12} \\
    \midrule
    Ours & \textbf{95.03} & \textbf{84.17} & \textbf{89.27} \\ 
       & \textcolor{blue}{+0.43} & \textcolor{blue}{+3.12} & \textcolor{blue}{+2.15} \\
    \bottomrule
    \end{tabular}
    }
    \end{subtable}
    \begin{subtable}[t]{.243\textwidth}
    \centering
    \caption{UCF101}
        \resizebox{0.99\linewidth}{!}{     \begin{tabular}{l cc|c}
    \toprule
    & Base & Novel & HM \\
    \midrule
    CLIP & 70.53 & {77.50} & 73.85 \\
    CoOp & {84.69} & 56.05 & 67.46 \\
    Co-CoOp & 82.33 & 73.45 & {77.64} \\  
    KgCoOp & 82.89 & 76.67  & 79.65 \\ 
    MaPLe & 83.00 &  {78.66} & {80.77} \\
    CoPrompt & \underline{86.90} & 79.57 & 83.07 \\
    CPL & 86.73 & \underline{80.17} & \underline{83.32} \\
    \midrule
    Ours & \textbf{88.30} & \textbf{83.05} & \textbf{85.60} \\ 
       & \textcolor{blue}{+1.40} & \textcolor{blue}{+2.88} & \textcolor{blue}{+2.28} \\
    \bottomrule
    \end{tabular}
    }
    \end{subtable}
\end{table*}

\paragraph{\textbf{Implementation Details.}}
To ensure a fair comparison, we employ the ViT-B/16 CLIP model across all three benchmark tasks. For base-to-novel generalization, 
we train our proposed CoCoLe with 16-shot images on base classes and subsequently evaluate it in both base classes and novel classes. 
For domain generalization and cross-dataset evaluation, we utilize the model trained with 16-shot ImageNet and test it on each target dataset.
Throughout the training, we keep both the visual and textual encoders fixed. Our data preprocessing follows CLIP's protocol, including resizing and random cropping operations, among others.
For the base-to-novel generation task, we conduct training for 30 epochs on ImageNet and 20 epochs on the other datasets.
$K_1$ and $K_2$ are set to 3 and 10 respectively. We set prompt length $M$ to 8, the size $N$ of the conceptual codebook to 100, and the number of selected concepts $K_3$ to 4.
The training is performed with a batch size of 8 with an initial learning rate of $10^{-3}$. We utilize the AdamW optimizer alongside a cosine annealing scheduler.

\subsection{Base-to-Novel Generalization}

In this section, We compare our CoCoLe method with seven baselines: zero-shot CLIP~\cite{radford2021learning}, CoOp~\cite{zhou2022learning}, CoCoOp~\cite{zhou2022conditional},  MaPLe~\cite{khattak2023maple}, KgCoOp~\cite{yao2023visual},
CoPrompt~\cite{roy2024consistency} and 
CPL~\cite{zhang2024concept}. \Cref{table:b_to_n} shows the experimental results for base-to-novel generalization across 11 datasets using 16-shot samples. 
We have bolded the top results and indicated improvements over the second-best performance in blue.
As we can see from Table \ref{table:average_acc}, the average of all 11 datasets shows that our method outperforms all the baselines by a large margin for both base and novel classes. In comparison to CoOp and CoCoOp, which are the pioneering prompt learning methods, the performance gain of our method even reached +11\% and +6.9\% respectively. Our method outperforms the previous state-of-the-art (CPL) by +2.28\% on novel classes and +1.62\% on the harmonic mean (HM). These results demonstrate the strong zero-shot generalization capability of our proposed method. Also, our method outperforms CPL on base classes by +0.84\%, which shows a strong few-shot learning capability. 

For the performance of individual datasets, our method outperforms CPL on all the datasets for novel class and HM. For the base classes, our method achieves superior performance gains compared to CPL on 8 out of 11 datasets. Even for Catech101, Flower102, and DTD, where there is a slight performance drop, it remains marginal. This highlights the enhanced generalization capability of our method towards novel classes without compromising performance on base classes.
Notably, aside from CPL, our method outperforms all other methods by a significant margin across all datasets. Compared to the second-best performing baseline, our method surpasses it by up to +3.32\%, +3.65\%, and +2.55\% on FGVCAircraft, DTD, and ImageNet, respectively.
These observations indicate that our method can effectively learn diverse and discriminative visual and textual concepts, thereby enhancing CLIP's adaptation for generalization tasks.

\subsection{Cross-Dataset Evaluation}
Table \ref{tab:crossdata} displays the comparison results with CoOp, CocoOp, MaPLe, CoPrompt, and CPL. Our CoCoLe approach achieves the top performance on both source and target datasets, averaging 68.91\% on the target set, surpassing CPL by 0.84\%. Notably, we observe the largest improvement of 2.26\% over CPL on UCF101, an action image dataset with distinct characteristics from ImageNet. This underscores that our method's conceptual codebook learning enhances generalization significantly.

\begin{table*}[t]
        \caption{Comparison with state-of-the-art methods on cross-dataset evaluation.}
        \centering
        
    \resizebox{\linewidth}{!}{
    \begin{tabular}{l c ccccccccccc}
    \toprule
    & \textbf{Source} & \multicolumn{11}{c}{\textbf{Target}} \\ \cmidrule(lr){2-2} \cmidrule(lr){3-13}
    & \rotbox{ImageNet} & \rotbox{Caltech101} & \rotbox{OxfordPets} & \rotbox{StanfordCars} & \rotbox{Flowers102} & \rotbox{Food101} & \rotbox{Aircraft} & \rotbox{SUN397} & \rotbox{DTD} & \rotbox{EuroSAT} & \rotbox{UCF101} & \rotbox{\emph{Average}} \\
    \midrule
    CoOp~\cite{zhou2022learning} & 71.51 & 93.70 & 89.14 & 64.51 & 68.71 & 85.30 & 18.47 & 64.15 & 41.92 & {46.39} & 66.55 & 63.88 \\
    CoCoOp~\cite{zhou2022conditional} & 71.02 &94.43 & 90.14 & 65.32 & 71.88 & 86.06 & 22.94 & 67.36 & 45.73 & 45.37 & 68.21 & 65.74 \\
    MaPLe~\cite{khattak2023maple} & 70.72 & 93.53 & 90.49 & 65.57 & 72.23 & 86.20 & 24.74 & 67.01 & 46.49 & 48.06 & 68.69 & 66.30 \\
    CoPrompt~\cite{roy2024consistency} & 70.80 & 94.50 & 90.73 & 65.67 & 72.30 & 86.43 & 24.00 & 67.57 & 47.07 & \textbf{51.90} & 69.73 & 67.00 \\
 CPL~\cite{zhang2024concept} & \underline{73.53} & \underline{95.52} & \underline{91.64} & \underline{66.17} & \underline{73.35} & \underline{87.68} & \underline{27.36} & \underline{68.24} & \underline{48.96} & 51.25 & \underline{70.52} & \underline{68.07} \\
  \textbf{Ours} & \textbf{73.88} & \textbf{95.88} & \textbf{91.93} & \textbf{67.79} & \textbf{74.17} & \textbf{87.97} & \textbf{28.83} & \textbf{68.75} & \textbf{49.26} & \underline{51.75} & \textbf{72.78} & \textbf{68.91} \\

    \bottomrule
    \end{tabular}
    }

    \label{tab:crossdata}
\end{table*}

\subsection{Domain Generalization}
We present the classification accuracy for both the source domain and target domains in Table~\ref{table:dg}, along with the average accuracy across the target domains.
Our approach consistently outperforms all baselines on both source and target datasets, setting a new state-of-the-art average accuracy of 61.85\% for this domain generalization task. Notably, our method beats CPL~\cite{zhang2024concept} by +1.51\% on ImageNet-R. This highlights the exceptional robustness of our model against distribution shifts.

\begin{table}[t]
\begin{center}
\caption{Comparison with existing methods on domain generalization task. The top results are highlighted in bold with the second-best results underlined.}
\resizebox{.6\linewidth}{!}{
\begin{tabular}{lcccccc}
\toprule
\multirow{2}*{Method}  & Source & \multicolumn{4}{c}{Target} \\ \cmidrule(lr){2-2} \cmidrule(lr){3-7} & ImageNet & -V2 & -Sketch & -A & -R & Ave. \\
\midrule
CLIP~\cite{radford2021learning} &  66.73 &  60.83  & 46.15  & 47.77 & 73.96 & 57.17 \\
CoOp~\cite{zhou2022learning}   & 71.51  & 64.20  & 47.99  & 49.71  & 75.21 & 59.28 \\
CoCoOp~\cite{zhou2022conditional}  & 71.02  & 64.07  & 48.75  & 50.63  & 76.18 & 59.90  \\
KgCoOp~\cite{yao2023visual}   & 71.20  & 64.10  & 48.97  & 50.69  & 76.70 & 60.11  \\
MaPLe~\cite{khattak2023maple}    & 70.72 & 64.07  & 49.15  & 50.90  & 76.98 & 60.27 \\
CoPrompt~\cite{roy2024consistency}  & 70.80  & 64.25  & 49.43  & 50.50  & \underline{77.51} & 60.42 \\
CPL~\cite{zhang2024concept}  & \underline{73.53}  & \underline{65.18}  & \underline{49.92}  & \underline{50.73}  & 77.38 & \underline{60.80} \\

\textbf{Ours}  & \textbf{73.88}  & \textbf{65.86}  & \textbf{50.89}  & \textbf{51.75}  & \textbf{78.89} & \textbf{61.85}\\
\bottomrule
\end{tabular}
}
\label{table:dg}
\end{center}
\end{table}

\subsection{Ablation Studies}
To rigorously assess our proposed approach, we conduct an empirical analysis of our design decisions and demonstrate the impact of various components in this section. Unless otherwise stated, our experiments are conducted on ImageNet for the Base-to-Novel Task, and we report the harmonic mean.

\paragraph{\textbf{Contributions of major algorithm components.}} In \cref{tab:ablation}, 
$\mathcal{L}_{ma}$ is used as a matching loss to optimize the visual concepts in the conceptual codebook. $\mathcal{L}_{cc}$ is for consistency constraint, while $\mathcal{L}_{or}$ is employed to orthogonalize the textual features of different prompts. We conduct an ablation study by systematically removing various components of our proposed CoCoLe to assess their individual importance. For evaluation, the first row of the table showcases the overall performance of CoCoLe, achieving a harmonic mean of 76.84\%. In the initial ablation experiment, we exclude $\mathcal{L}{or}$ from CoCoLe, resulting in a performance decrease of 0.46\%. This underscores the significance of $\mathcal{L}{or}$ in CoCoLe.
Next, we remove $\mathcal{L}_{cc}$, effectively enforcing consistency between the learnable conceptual prompts and handcrafted concept-based prompts. As a result, there is a decrease in performance by 1.98\%, indicating the significance of the regularization strategy.
Finally, we remove $\mathcal{L}_{ma}$, which leads to a performance drop of 2.39\%. This shows that the learnable conceptual codebook plays a crucial role in CoCoLe. In general, the overall results highlight the significant contribution of all components to enhanced performance.

\begin{figure}[t]
    \begin{minipage}[c]{0.49\linewidth}
        \captionof{table}{The ablation study on each component of CoCoLe. }
        \label{tab:ablation}
        \centering
        \setlength{\tabcolsep}{1mm}
            \scalebox{0.9}{
            \begin{tabular}{ccc|c} 
            \toprule
              $\mathcal{L}_{ma}$   & $\mathcal{L}_{cc}$ & $\mathcal{L}_{or}$ &  Accuracy(HM)  \\ 
            \midrule
            \cmark &\cmark &\cmark  &76.84 \\
            \cmark &\cmark &\xmark  &76.38 \\%
            \cmark &\xmark &\cmark  &74.86 \\
            \xmark &\cmark &\cmark  &74.45 \\%
            \cmark &\xmark &\xmark &74.57 \\
            \xmark &\cmark &\xmark  &74.26 \\%
            \xmark &\xmark &\xmark  &72.12  \\%
            \bottomrule
            \end{tabular}}
\end{minipage}
\hfill
\begin{minipage}[c]{0.49\linewidth}
\centering
\captionof{table}{Ablation study of  $M$, $N$, and $K_3$. }
\label{table:conceptnum}
\setlength{\tabcolsep}{0.6mm}
    \scalebox{0.9}{
\begin{tabular}{c|ccccc}
\toprule
\textbf{Value of $M$}  & 4   & 
\textbf{8} & 12  & 16  &20\\ \midrule
\textbf{Accuracy}  & 75.83 & \textbf{76.84} & 76.53 & 75.97 & 75.75  \\ \midrule
\textbf{Value of $N$}  & 50   & \textbf{100} & 150 & 200  &250 \\ \midrule
\textbf{Accuracy}  & 75.07 & \textbf{76.84} & {76.73} & 76.51 & 76.25 \\ \midrule
\textbf{Value of $K_3$}  & 1   & 2 & \textbf{4} & 6 & 8 \\ \midrule
\textbf{Accuracy}  & 75.30 & 75.86 & \textbf{76.84} & 76.72 & 76.33 \\ 
\bottomrule
\end{tabular}}
\end{minipage}
\end{figure}

\paragraph{\textbf{The value of $M$, $N$, and $K_3$.}}
As defined in \cref{sec:ccl}, $M$ is the length of prompts, $N$ is the size of the conceptual codebook, and $K_3$ is the number of learned visual concepts (\ie, keys in the codebook) selected for training at the same time. As indicated in Table \ref{table:conceptnum}, the model performs optimally with $M=8$. If the prompt is overly long, it raises both training time and computational costs. Concerning $N$, the model excels with $N=100$. Additionally, experimenting with various $K_3$ values shows that saturating training with excessive keys and prompts simultaneously diminishes model performance. The model achieves its peak performance at $K_3=4$.

\paragraph{\textbf{Comparison on the training time.}} As shown in \cref{tab:time}, our proposed CoCoLe exhibits a significant performance advantage over other methods. Though our CoCoLe requires more training time compared to CoOp, it still outperforms CoOp by 11\%. Furthermore, when compared to CoCoOp and CPL, our method achieves substantial performance gains with less time. This showcases the efficiency and effectiveness of our proposed CoCoLe.

\begin{table}[t]
\caption{Comparison on the training time. We report the average accuracy across 11 datasets on base-to-novel tasks.}
\label{tab:time}
\centering
\footnotesize
\begin{tabular}{l|c|ccc|c}
\toprule
\multirow{2}{*}{Methods} & \multirow{2}{*}{Prompts} & \multicolumn{3}{c|}{Accuracy} &\multirow{2}{*}{Training-time} \\ \cline{3-5}
                         &                            & Base       & New       & H      &      \\

\midrule
CLIP                     & handcrafted &  69.34 & 74.22  &  71.70 & - \\
\midrule
CoOp                     & textual& 82.69&63.22 & 71.66 &  6ms/image                   \\
CoCoOp                   &textual+visual& 80.47& 71.69 & 75.83 & 160ms/image              \\
CPL                  & textual+visual&  84.38&   78.03& 81.08& 25ms/image                \\
\textbf{CoCoLe}                      &textual+visual &  \textbf{85.22}&   \textbf{80.31} & \textbf{82.70} &  \textbf{10ms}/image              \\
\bottomrule
\end{tabular}
\end{table}

\paragraph{\textbf{Visualization. }} As shown in Fig.~\ref{fig:visual}, in order to confirm that various prompts indeed capture distinct image concepts, we employ Grad-CAM~\cite{selvaraju2017grad} to visually represent the image contents associated with different prompts. 
Observing \cref{fig:visual}(a), it's apparent that various prompts highlight distinct regions within the same image, showcasing the diversity of the acquired prompts. For instance, different prompts applied to the Koala image emphasize different areas such as the head, claws, and tree, illustrating the versatility of the learned prompts.
To determine if the learned prompts indeed encapsulate higher-level semantics image concepts, we visualize the content of specific prompts ($\textbf{P}_3$, $\textbf{P}_{23}$, $\textbf{P}_{48}$) across different images in \cref{fig:visual}(b). It's evident that $\textbf{P}_3$ mainly captures the concept "wheels," $\textbf{P}_{23}$ encapsulates the concept "grass," while $\textbf{P}_{48}$ focuses on the "screen" of the devices. This highlights the effectiveness of prompts in learning key concepts that can be generalized across images, thereby enhancing performance in generalization tasks. In the Supplemental Materials, we provide additional details of our proposed CoCoLe and experimental results.

\section{Conclusion}
To address the problem of enhancing the generalization capability of VLMs while adapting them to downstream tasks, we propose Conceptual Codebook Learning (CoCoLe). The learned conceptual codebook consists of visual concepts as keys and conceptual prompts as values, which serve as a link between the image encoder's outputs and the text encoder's inputs. Additionally, we incorporate a handcrafted concept cache as a regularization to alleviate the overfitting issues in low-shot scenarios. Extensive experimental results demonstrate that our CoCoLe method remarkably outperforms the existing state-of-the-art methods.

\section*{Acknowledgements}
 This research was supported by the National Natural Science Foundation of
China (No. 62331014) and Project 2021JC02X103.

\bibliographystyle{splncs04}
\bibliography{main}
\end{document}


\title{Supplemental Materials}
\maketitle

\section{Relationship with Existing Methods}
\begin{figure}[h]
    \centering
    \includegraphics[width=\linewidth]{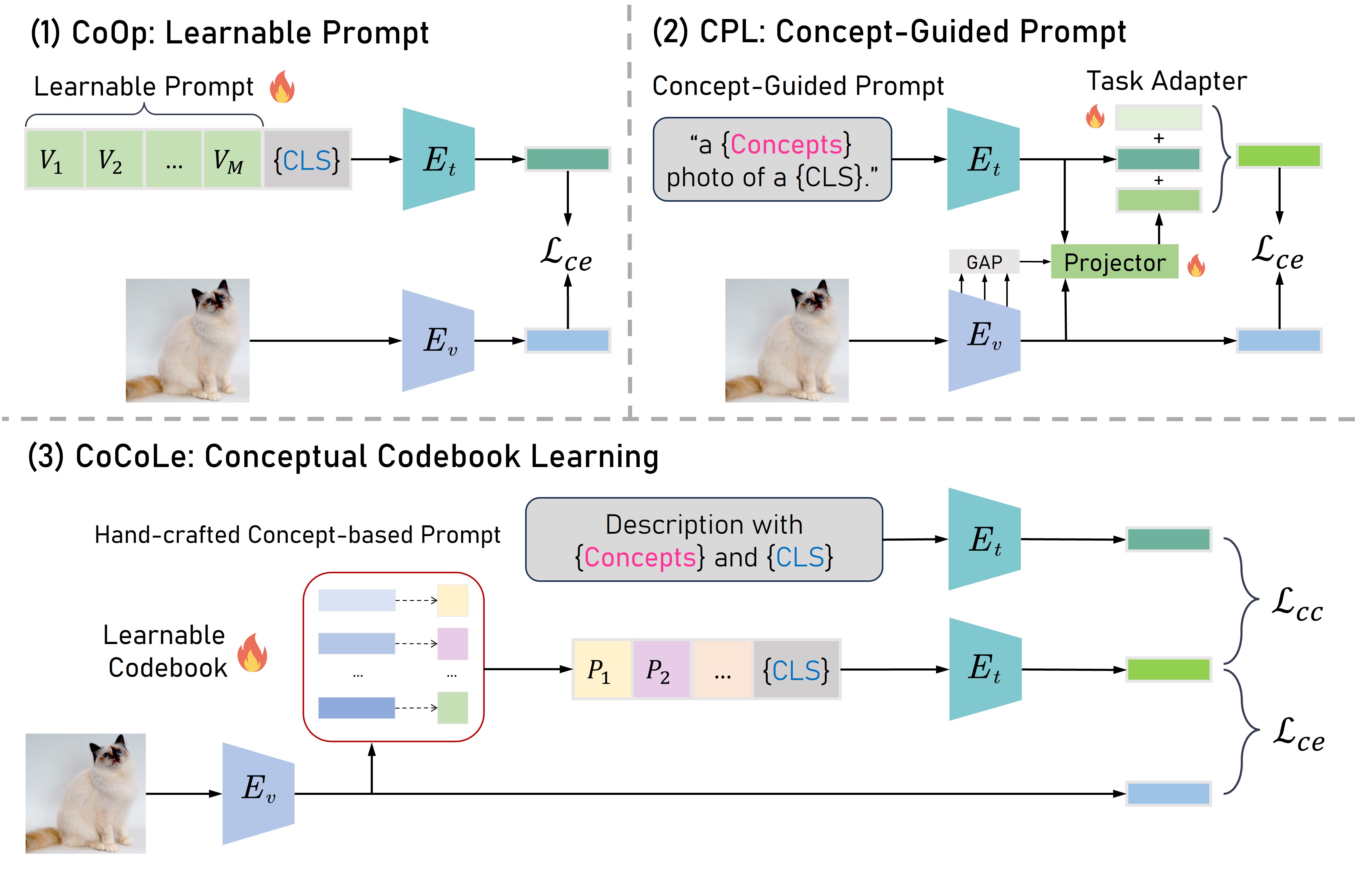}
    \caption{Comparison between related baseline methods and our proposed CoCoLe.}
    \label{fig:com}
    \vspace{-5pt}
\end{figure}

\Cref{fig:com} shows the baseline methods that are closely related to our methods. CoCoLe replaces the specific prompts of CoOp~\cite{zhou2022learning} with a learnable codebook to store the conceptual prompts in key-value pairs. Being regularized by  hand-crafted concept-based prompts, CoCoLe achieves better generalization capabilities while learning adaptive prompts when compared to the CPL method~\cite{zhang2024concept}.

\section{Additional Experimental Results}

\subsection{Dataset Details}

In \Cref{tab:dataset}, we follow the few-shot evaluation protocol to evaluate our method on 11 image classification datasets, including generic object classification (ImageNet~\cite{recht2019imagenet}, Caltech101~\cite{fei2004learning}), fine-grained object classification (OxfordPets~\cite{parkhi2012cats}, StandfordCars~\cite{krause20133d}, Flowers102~\cite{nilsback2008automated}, Food-101~\cite{bossard2014food}, FGVCircraft~\cite{maji2013fine}), texture classification (DTD~\cite{cimpoi2014describing}), remote sensing recognition (EuroSAT~\cite{helber2019eurosat}), scene recognition (SUN397~\cite{xiao2010sun}), and action recognition (UCF101~\cite{soomro2012ucf101}). We evaluate the domain generalization performance on ImageNet~\cite{deng2009imagenet}, ImageNet-V2~\cite{recht2019imagenet} and ImageNet-Sketch~\cite{wang2019learning}.
The details of each dataset are shown in \Cref{tab:dataset}, including the number of classes, the sizes of training and testing sets, and the task descriptions.

\begin{table*}[t]
\vspace{-5pt}
\centering
    \caption{The detailed statistics of datasets used in experiments. }
    \label{tab:dataset}
    \resizebox{.8\textwidth}{!}{
    \small
    \begin{tabular}{lcccc}
    \toprule
Dataset                   &  Classes   &  Training size   &  Testing size  &  Task description \\ \midrule
ImageNet~\cite{recht2019imagenet}  &  1,000  &  1.28M  &  50,000  &  Object recognition \\
Caltech101~\cite{fei2004learning}  &  100  &  4,128  &  2,465 &  Object recognition \\
OxfordPets~\cite{parkhi2012cats}  &  37   &  2,944  &  3,669  &  Fine-grained pets recognition \\ 
StanfordCars~\cite{krause20133d}  &  196  &  6,509  &  8,041  &  Fine-grained car recognition \\
Flowers102~\cite{nilsback2008automated}  &  102  &  4,093  &  2,463  &  Fine-grained flowers recognition \\ 
Food101~\cite{bossard2014food}  &  101  &  50,500 &  30,300  &  Fine-grained food recognition  \\ 
FGVCAircraft~\cite{maji2013fine}  &  100  &  3,334  &  3,333  &  Fine-grained aircraft recognition\\
SUN397~\cite{xiao2010sun} &  397 &  15,880  &  19,850  &  Scene recognition\\ 
DTD~\cite{cimpoi2014describing} &  47  &  2,820  &  1,692  &   Texture recognition\\ 
EuroSAT~\cite{helber2019eurosat} &  10  &  13,500  &  8,100  &  Satellite image recognition \\ 
UCF101~\cite{soomro2012ucf101} &  101  &  7,639  &  3,783  &  Action recognition\\
\midrule
ImageNet-V2~\cite{recht2019imagenet}  &  1,000  &  -  &  10,000  &  Robustness of collocation  \\
ImageNet-Sketch~\cite{wang2019learning}  &  1,000  &  -  & 50,889  &  Robustness of sketch domain\\
ImageNet-A~\cite{hendrycks2021natural} &  200  &  -  & 7,500  & Robustness of adversarial attack\\
ImageNet-R~\cite{hendrycks2021many} &  200  &  -  & 30,000 & Robustness of multi-domains\\
   \bottomrule
    \end{tabular}
    }
\end{table*}

\subsection{Text Concept Details}
In Sec. 3.2, we constructed a list $\Omega_t$ comprising a set of descriptive text concepts. Here, we provide more examples of the text concepts in \Cref{tab:textcon}.

\begin{table}[t]
    \centering
    
    \caption{More examples of the text concepts in the handcrafted concept cache.}
    \label{tab:textcon}
    \scalebox{0.8}{
    \begin{tabular}{c|c}
    \toprule
    Class & Text Concept Examples\\
    \midrule
    Texture    &  \texttt{smooth, rough, grainy, wrinkled, bumpy, fuzzy, polished, hairy} \\
    Color     &  \texttt{white, black, blue, yellow, gold, green, red, pink, magenta} \\
    Transparency & \texttt{clear, opaque, translucent, frosted, glassy, sheer, film, hazy} \\
    Brightness & \texttt{bright, glowing, dull, radiant, luminous, gleaming, glaring} \\
    Motion & \texttt{fast, still, moving, vibrating, swaying, rotating, fluttering} \\
    Emotion & \texttt{sadness, joy, confusion, disgust, surprise, anger, excitement} \\
    Pattern & \texttt{striped, geometric, floral, damask, tartan, chekered, Herringbone} \\
    \bottomrule
     \end{tabular}
    }

\end{table}

\subsection{Additional Ablation Studies}
In Sec. 4.5, we provided an ablation study on the contributions of different components of our method. Here, we provide additional ablation studies on the impact of similarity metrics and the selection of $K_1$ and $K_2$ parameters in our method.

\begin{figure}[t]
    \begin{minipage}[c]{0.49\linewidth}
        \captionof{table}{Ablation study on similarity metrics of loss function used by CoCoLe. All experiments are conducted on ImageNet of the base-to-novel task. $cos$ for using cosine similarity and $eu$ for using Euclidean distance.}
        \vspace{10pt}
        \label{tab:metric}
        \centering
        \setlength{\tabcolsep}{1mm}
            \scalebox{0.8}{
            \begin{tabular}{cc|c} 
            \toprule
              $\mathcal{L}_{ma}$   & $\mathcal{L}_{cc}$ &   Accuracy(HM)  \\ 
            \midrule
            $cos$ & $eu$  &\textbf{76.84} \\
            $cos$ & $cos$  &76.52 \\%
            $eu$ & $cos$  &76.27\\%
            $eu$ &  $eu$  &76.40 \\
            \bottomrule
            \end{tabular}}
\end{minipage}
\hfill
\begin{minipage}[c]{0.49\linewidth}
\centering
\captionof{table}{Ablation study of the number $K_1$ of selected image features for the key of the handcrafted concept cache, the number $K_2$ of selected visual concepts from the cache at a time. }
\label{tab:k12}
\vspace{16pt}
\setlength{\tabcolsep}{0.4mm}
    \scalebox{0.86}{
\begin{tabular}{c|ccccc}
\toprule
\textbf{Value of $K_1$}  &  1   & 
\textbf{3} & 5  & 8  &10\\ \midrule
\textbf{Accuracy}  & 76.51 & \textbf{76.84} & 76.32 & 75.91 & 75.75  \\ \midrule
\textbf{Value of $K_2$}  & 1   & 3  &  5 & 10  &15 \\ \midrule
\textbf{Accuracy}  & 75.07 & 75.80 & 76.38 & \textbf{76.84} & 76.12 \\   
\bottomrule
\end{tabular}}
\end{minipage}
\end{figure}

\paragraph{\textbf{(1) Selection of Similarity Metrics.}} In \Cref{tab:metric}, $\mathcal{L}_{ma}$ is the matching loss used to optimize the keys, $\mathcal{L}_{cc}$ is for the consistency constraint. In this ablation study, we select different metrics for these loss functions including cosine similarity and Euclidean distance. From  \Cref{tab:metric}, we can see that, for the matching loss, cosine similarity exhibits better performance than the Euclidean distance loss. On the other hand, for consistency loss, an Euclidean is much better at enforcing strong regularization textual output to guarantee generalization.

\paragraph{\textbf{(2) The value of $K_1$ and $K_2$.}} Besides $M$, $N$, and $K_3$, we conduct more ablation studies on the number of selected image features for the key of the handcrafted concept cache ($K_1$) and the number of selected visual concepts from the cache a time ($K_2$). As shown in \Cref{tab:k12}, the model achieves the optimal performance with $K_1=3$, because a larger number of $K_1$ might introduce additional noise to the keys, leading to the discrepancy between modalities. As to $K_2$, we set it as $10$, because it is necessary to enrich the textual input with comprehensive prompts so that the model can maintain robustness against domain shift.

\begin{table}[t]
        \captionof{table}{Ablation study on the large language model that used by CoCoLe.}
        \vspace{10pt}
        \label{tab:llm}
        \centering
        \setlength{\tabcolsep}{1mm}
            \scalebox{0.8}{
            \begin{tabular}{c|c} 
            \toprule
              Type   & Accuracy(HM)  \\ 
            \midrule
            w/ LLM &\textbf{76.84} \\
            w/o LLM   &76.20 \\
            \bottomrule
            \end{tabular}
            }
\end{table}

\paragraph{\textbf{(3) Ablation studies on the large language model.}} In Section 3.2, we utilize a large language model to orgainze handcrafted concept-based prompts. Here we conduct an ablation study on the LLM to generate handcrafted concept-guided prompts. From \Cref{tab:llm}, we can see that the generated prompts are better than random arrangement of selected text concept, which could stem from LLM's ability to generate a more grammatical and logical description.

\bibliographystyle{splncs04}
\bibliography{main}